\RequirePackage{snapshot}
\documentclass{article}

\usepackage[nonatbib, final]{nips_2017}
\usepackage[square,numbers,sort&compress]{natbib}

\usepackage[utf8]{inputenc} \usepackage[T1]{fontenc}    \usepackage{hyperref}       \hypersetup {
  colorlinks = True,
}
\usepackage{url}            \usepackage{booktabs}       \usepackage{amsfonts}       \usepackage{nicefrac}       \usepackage{microtype}      
\usepackage{graphicx}
\usepackage{amsmath}
\usepackage{amssymb}
\usepackage{amsthm}
\usepackage[font=small]{caption}
\usepackage{subcaption}

\usepackage{tablefootnote}
\usepackage{adjustbox}

\newcommand{\ignore}[1]{}

\usepackage{color}

\newif\ifkeepComments
\keepCommentstrue

\newcommand{\methodTag}[0]{Attentional Pooling}
\newcommand{\tableSize}[0]{\small}
\newcommand{\customfootnotesize}[0]{\scriptsize}

\title{Attentional Pooling for Action Recognition}

\author{
  Rohit Girdhar \qquad Deva Ramanan \\
  The Robotics Institute, Carnegie Mellon University \\
    \footnotesize{\url{http://rohitgirdhar.github.io/AttentionalPoolingAction}}
                                          }

\begin{document}

\maketitle

\begin{abstract}
We introduce a simple yet surprisingly powerful
model to incorporate attention in action recognition and human
object interaction tasks. Our proposed attention module can be
trained with or without extra supervision, and gives a sizable boost
in accuracy
while keeping the network size 
and computational cost nearly the same.
It leads to significant improvements over state of the art
base architecture on three standard action recognition benchmarks across
still images and videos, and establishes new state of the art
on MPII dataset with 12.5\% relative improvement.
We also perform an extensive analysis of our attention
module both empirically and analytically. In terms of the latter, we introduce a novel derivation of bottom-up and top-down attention as low-rank approximations of bilinear pooling methods (typically used for fine-grained classification). From this perspective, our attention formulation suggests a novel characterization of action recognition as a fine-grained recognition problem.
\end{abstract}

 \section{Introduction}

Human action recognition is a fundamental and well studied
problem in computer vision.
Traditional approaches to action
recognition relied on object detection~\cite{desai10action,gupta09pami,yao10actions2},
articulated
pose~\cite{maji2011action,pishchulin14gcpr,ramanan2003automatic,yang10poseaction,yao10actions2},
dense trajectories~\cite{WangCVPR11,IDT_Wang_13} and part-based/structured
models~\cite{Delaitre10,yao10actions,yao11actions}.
However, more recently these methods have been surpassed by deep CNN-based
representations~\cite{gkioxari15rstar,mallya16actions,Simonyan_14b,Tran_15}.
Interestingly, even video based action recognition has benefited
greatly from advancements in image-based CNN models~\cite{He_16,Ioffe_15,Simonyan_14a,Szegedy_16}.
With the exception of a few 3D-conv-based methods~\cite{Tran_15,Varol_16,piergiovanni2017learning},
most approaches~\cite{LRCN,Feichtenhofer_16b,Feichtenhofer_16,Girdhar_17a_ActionVLAD,WangL_16a},
including the current state of the art~\cite{WangL_16a}, use a variant of
discriminatively trained 2D-CNN~\cite{Ioffe_15}
over the appearance (frames) and in some cases motion (optical flow) modalities of the input video.

{\bf Attention:} While using standard deep networks over the full image have shown great promise
for the task~\cite{WangL_16a}, it raises the question of whether
action recognition can be considered as a general classification problem.
Some recent works have tried to generate more {\em fine-grained
} representations
by extracting features around human pose keypoints~\cite{cheronICCV15}
or on object/person bounding 
boxes~\cite{gkioxari15rstar,mallya16actions}. This form of `hard-coded attention'
helps improve performance, but requires labeling (or detecting) objects  
or human pose. Moreover,
these methods assume that focusing on the human or its parts is always useful
for discriminating actions. This might not necessarily be true for all actions; 
some actions might be easier to distinguish using the background and context, like 
a `basketball shoot' vs a `throw'; while others might require paying close
attention to objects being interacted by the human, like in case of 
`drinking from mug' vs `drinking from water bottle'.

{\bf Our work:} In this work, we propose a simple yet surprisingly powerful
network modification that learns attention maps which focus computation on specific parts of the input relevant to the task at hand. Our attention maps can be
learned without any additional supervision and automatically lead to significant
improvements over the baseline architecture. Our formulation is simple-to-implement, and can be seen as a natural extension of average pooling into a ``weighted'' average pooling with image-specific weights. Our formulation also provides a novel factorization of attentional processing into bottom-up saliency combined with top-down attention. We further experiment with adding human pose as an intermediate supervision to train our attention module, which encourages the network to look for human object interactions. While this makes little difference to the performance
of image-based recognition models, it leads to a larger improvement on video
datasets as videos consist of large number of `non-iconic' frames where the
subject of object of actions may not be at the center of focus.

{\bf Our contributions:} (1) An easy-to-use
extension of state-of-the-art base
architectures
that incorporates attention to give significant improvement in
action recognition performance at virtually negligible increase
in computation cost;
(2) Extensive analysis of its performance on three 
action recognition datasets across still images and videos,
obtaining state of the art on MPII and HMDB-51 (RGB, single-frame models) and competitive on HICO;
(3) Analysis of different base architectures for applicability
of our attention module; and
(4) Mathematical analysis of our proposed attention module and showing
its equivalence to a rank-1 approximation of second order 
or bilinear pooling (typically used in fine grained
recognition methods~\cite{gao16cbp,kong2016low,lin2015bilinear}) suggesting
a novel characterization of action recognition as a fine grained
recognition problem.

 \section{Related Work}

Human action recognition is a well studied problem with
various standard benchmarks spanning across
still images~\cite{chao15hico,Everingham2010,pishchulin14gcpr,ronchi15cocoa,yao11actions} and
videos~\cite{kay2017kinetics,hmdb51,charades,ucf101}. The newer image based datasets
such as HICO~\cite{chao15hico} and MPII~\cite{pishchulin14gcpr}
are large and highly diverse, containing 600 and 393 classes
respectively. In contrast, collecting such diverse video
based action datasets is hard, and hence existing popular benchmarks like
UCF101~\cite{ucf101} or HMDB51~\cite{hmdb51} contain only
101 and 51 categories each. This in turn has lead to much higher
baseline performance on videos, eg.\ $\sim 94\%$~\cite{WangL_16a} classification accuracy on UCF101,
compared to images, eg.\ $\sim 32\%$~\cite{mallya16actions} mean average precision (mAP) on MPII.

\noindent {\bf  Features:} Video based action recognition methods focus on two main
problems: action classification and (spatio-)temporal detection.
While image based recognition problems, including action
recognition, have seen a large
boost with the recent advancements in deep learning
(e.g., MPII performance went up from 5\% mAP~\cite{pishchulin14gcpr}
to 27\% mAP~\cite{gkioxari15rstar}),
video based recognition still relies on
hand crafted features such as iDT~\cite{IDT_Wang_13} to
obtain competitive performance. These features are computed 
by extracting
appearance and motion features along densely sampled point
trajectories in the video, aggregated into a fixed
length representation by using fisher vectors~\cite{Perronnin_07}.
Convolutional neural network (CNN) based approaches
to video action recognition have broadly followed two main
paradigms: (1) Multi-stream networks~\cite{Simonyan_14b,WangL_16a}
which split the input video into multiple modalities such
as RGB, optical flow, warped flow etc, train standard
image based CNNs on top of those, and late-fuse the predictions from
each of the CNNs; and (2) 3D Conv Networks~\cite{Tran_15,Varol_16}
which represent the video as a spatio-temporal blob and train a 3D
convolutional model for action prediction. In terms of performance,
3D conv based methods have been harder to scale and
multi-stream methods~\cite{WangL_16a}
currently hold state of the art performance
on standard benchmarks. Our approach is complementary to
these paradigms and the attention module can be applied
on top of either. We show results on improving action
classification over state of the art multi-stream model~\cite{WangL_16a}
in experiments. 
\noindent {\bf Pose:} There have also been previous works in incorporating human
pose into action recognition~\cite{cheronICCV15,delaitre11hoi,zolf17chained}.
In particular, P-CNN~\cite{cheronICCV15}
computes local appearance and motion features along the pose
keypoints and aggregates those over the video for action 
prediction, but is not end-to-end trainable.
More recent work~\cite{zolf17chained} adds pose
as an additional stream in chained multi-stream fashion and shows
significant improvements.
Our approach is complementary to these approaches as we use pose as a
regularizer in learning spatial attention maps to weight
regions of the RGB frame. Moreover, our method is not constrained
by pose labels, and as we show in experiments, can show effective performance with pose predicted by
existing methods~\cite{cao2017realtime} or even without using pose.

\noindent {\bf Hard attention:} Previous works in image based action recognition have shown impressive
performance by incorporating evidence from the human,
context and pose keypoint bounding boxes~\cite{cheronICCV15,gkioxari15rstar,mallya16actions}.
Gkioxari {\it el al.}~\cite{gkioxari15rstar} modified R-CNN pipeline
to propose R*CNN, where they choose an auxiliary box to encode
context apart from the human bounding box.
Mallya and Lazebnik~\cite{mallya16actions} improve upon it
by using the full image as the context and using multiple
instance learning (MIL) to reason over all humans present in the image
to predict an action label for the image. Our approach
gets rid of the bounding box detection step and improves
over both these methods by automatically learning to attend
to the most informative parts of the image for the task.

\noindent {\bf Soft attention:} There has been relatively little work that explores
unconstrained `soft' attention for
action recognition, with the exception of~\cite{sharma2015attention,song2016end}
for spatio-temporal and~\cite{shi2016joint} for temporal attention.
Importantly, all these consider a video setting, where a LSTM network 
predicts a spatial attention map for the current frame. Our method, however, uses a single frame to both predict and apply spatial attention, making it amenable to both single image and video based use cases. 
\cite{song2016end} also uses
pose keypoints labeled in 3D videos to drive attention to parts of the body. In contrast, we
learn
an unconstrained attention model that frequently learns to look around the human body for objects
that make it easier to classify the action.

\noindent {\bf Second-order pooling:}  Because our model uses a single set of appearance features to both predict and apply an attention map, this makes the output {\em quadratic} in the features (Sec.~\ref{sec:att-pool-bilinear}). This observation allows us to implement attention through second-order or bilinear pooling operations~\cite{lin2015bilinear}, made efficient through low-rank approximations~\cite{gao16cbp,kim2016hadamard,kong2016low}.  Our work is most related to \cite{kong2016low}, who point out when efficiently implemented, low-rank approximations avoid explicitly computing second-order features. We point out that a rank-1 approximation of second-order features is equivalent to an attentional model sometimes denoted as ``self attention''~\cite{vaswani2017attention}. Exposing this connection allows us to explore several extensions, including variations of bottom-up and top-down attention, as well as regularized attention maps that make use of additional supervised pose labels.

 \section{Approach}

Our attentional pooling module is a trainable layer that plugs in as a replacement for a pooling operation in any standard CNN. As most contemporary architectures~\cite{He_16,Ioffe_15,Szegedy_16}
are fully convolutional with an average pooling operation at the end, our
module can be used to replace that operation with an attention-weighted pooling.
We now derive the pooling layer as an efficient low-rank approximation to second
order pooling (Sec.~\ref{sec:att-pool-bilinear}). Then,
we describe our network architecture that incorporates this attention module and explore
a pose-regularized variant of the same (Sec.~\ref{sec:att-pool-bilinear-nw}).

\subsection{Attentional pooling as low-rank approximation of second-order pooling}\label{sec:att-pool-bilinear}

Let us write the layer to be pooled as $X \in R^{n \times f}$, where $n$ is the number of spatial locations (e.g., $n = 16 \times 16=256$) and $f$ is the number of channels (e.g., $2048$). Standard sum (or max) pooling would reduce this to vector in $R^{f \times 1}$, which could then be processed by a ``fully-connected'' weight vector ${\bf w} \in R^{f \times 1}$ to generate a classification score. We will denote matrices with upper case letters, and vectors with lower-case bold letters. For the moment, assume we are training a binary classifier (we generalize to more classes later in the derivation). We can formalize this pipeline with the following notation:
\begin{align}
  score_{pool}(X) =  {\bf 1}^T X {\bf w}, \qquad \text{where} \qquad X \in R^{n \times f}, {\bf 1} \in R^{n \times 1}, {\bf w} \in R^ {f \times 1} \label{eq:pool}
\end{align}

\noindent where $ {\bf 1} $ is a vector of all ones and ${\bf x} =  {\bf 1}^T X \in R^{1 \times f}$ is the (transposed) sum-pooled feature.

{\bf Second-order pooling:} Following past work on second-order pooling~\cite{carreira2012semantic}, let us construct the feature $X^TX \in R^{f \times f}$. Prior work has demonstrated that such second-order statistics can be useful for fine-grained classification~\cite{lin2015bilinear}. Typically, one then ``vectorizes'' this feature, and learns a $f^2$ vector of weights to generate a score. If we write the vector of weights as a $f \times f$ matrix, the inner product between the two vectorized quantities can be succinctly written using the trace operator\footnote{\customfootnotesize \url{https://en.wikipedia.org/wiki/Trace_(linear_algebra)}}.
The key identity, $Tr(AB^T) = dot(A(:),B(:))$ (using matlab notation), can easily be verified by plugging in the definition of a trace operator. This allows us to write the classification score as follows:
\begin{align}
 score_{order2}(X) =  Tr(X^TX W^T), \qquad \text{where} \qquad X \in R^{n \times f}, W \in R^ {f \times f} \label{eq:order2}
\end{align}

{\bf Low-rank second-order pooling:} Let us approximate matrix $W$ with a rank-1 approximation, $W={\bf a}{\bf b}^T$ where ${\bf a},{\bf b}\in R^{f \times 1}$. Plugging this into the above yields a novel formulation of attentional pooling:
\begin{align}
  score_{attention}(X) &=  Tr(X^TX {\bf b a}^T), \qquad \text{where} \qquad X \in R^{n \times f}, {\bf a,b} \in R^{f \times 1}\\
&=  Tr({\bf a}^T X^T X {\bf b}) \label{eq:cycle}\\
&= {\bf a}^T X^T X {\bf b} \label{eq:scalar}\\
&= {\bf a}^T \Big( X^T (X {\bf b}) \Big) \label{eq:final}
\end{align}
where \eqref{eq:cycle} makes use of the trace identity that $Tr(ABC) = Tr(CAB)$ and \eqref{eq:scalar} uses the fact that the trace of a scalar is simply the scalar. The last line \eqref{eq:final} gives efficient implementation of attentional pooling: given a feature map $X$, compute an attention map over all $n$ spatial locations with ${\bf h} = X{\bf b} \in R^{n \times 1}$, that is then used to compute a weighted average of features ${\bf x} = X^T {\bf h} \in R^{f \times 1}$. This weighted-average feature is then pushed through a linear model ${\bf a}^T {\bf x}$ to produce the final score.

\begin{figure}
    \centering
    \begin{subfigure}{.68\textwidth}
      \centering
      \includegraphics[width=\linewidth]{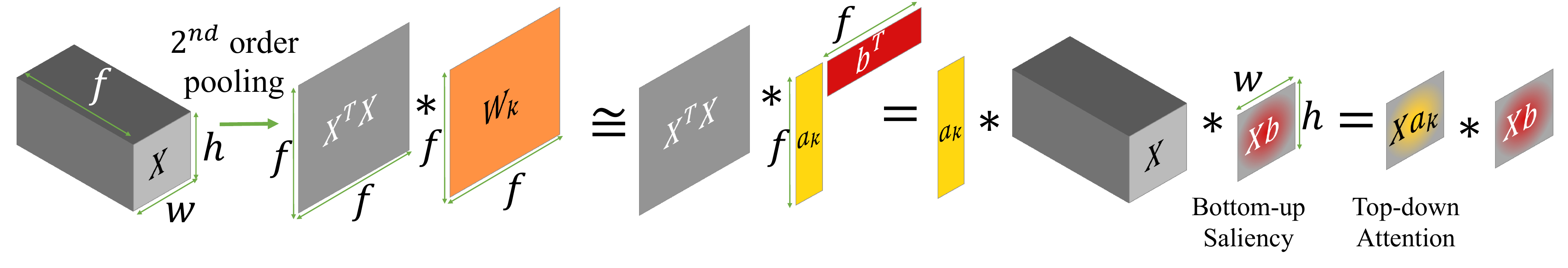}
      \caption{Visualization of our approach to attentional pooling as a rank-1
approximation of $2^{nd}$ order pooling.
By judicious ordering of the matrix multiplications, one can avoid computing the second order feature $X^TX$ and instead compute the product of two attention maps. The top-down attentional map is computed using class-specific weights $a_k$, while the bottom-up map is computed using class-agnostic weights $b$. We visualize the top-down and bottom-up
attention maps learned by our approach in Fig.~\ref{fig:mpii-heatmaps}.}
      \label{fig:nwarch-math}
    \end{subfigure}\hfill
   \begin{subfigure}{.3\textwidth}
      \centering
      \includegraphics[width=\linewidth]{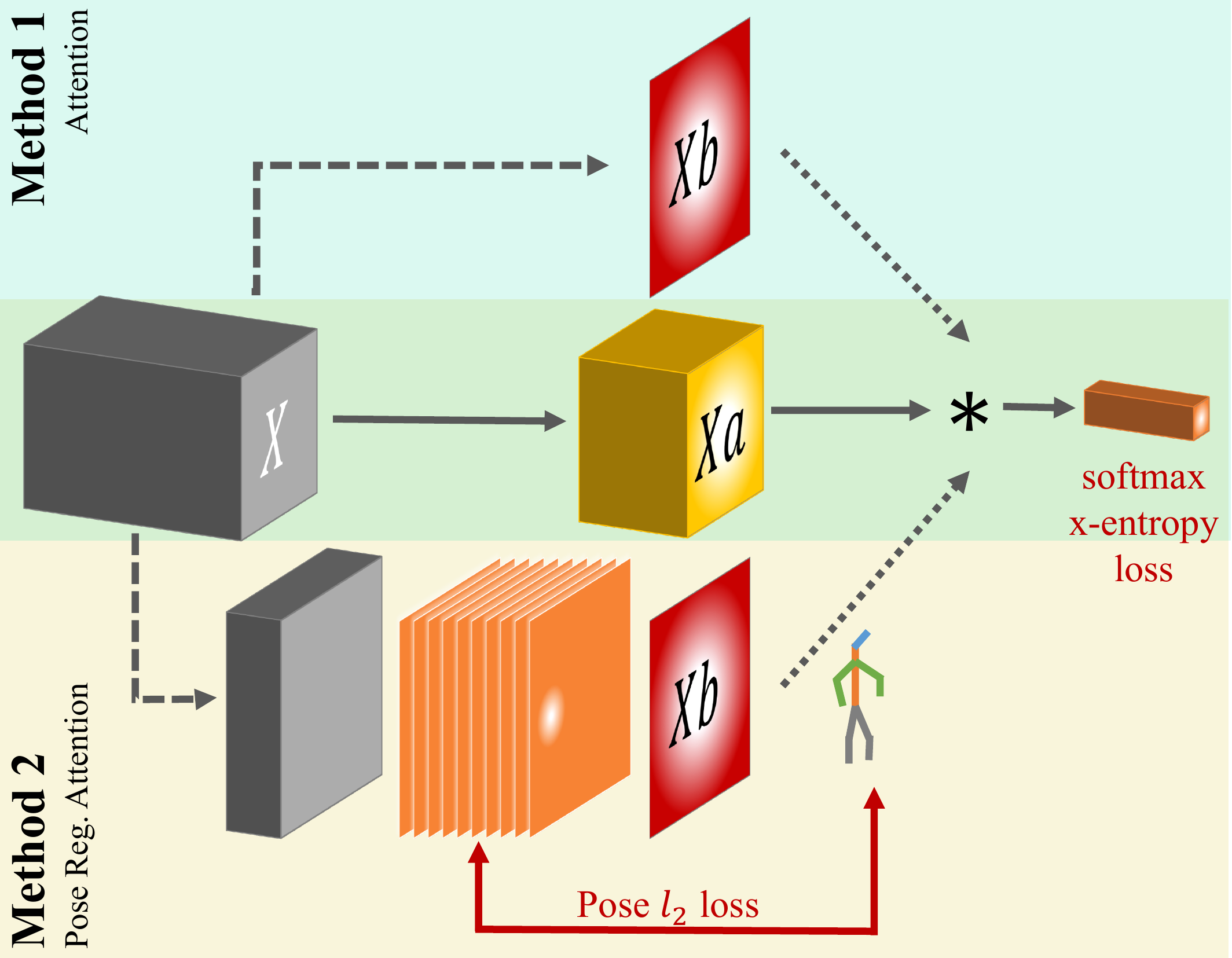}
      \caption{We explore two architectures in our work, explained in Sec.~\ref{sec:att-pool-bilinear-nw}.}
      \label{fig:nwarch}
   \end{subfigure}
    \caption{Visualization of our derivation and final network architectures.}
    \label{fig:nwarch-combined}
\end{figure}

Interestingly, \eqref{eq:final} can also be written as the following:
\begin{align}
  score_{attention}(X)  &= \Big( (X  {\bf a})^T X \Big) {\bf b}\\
&=  (X  {\bf a})^T (X {\bf b})
\end{align}
The first line illustrates that the attentional heatmap can also be seen as $ X{\bf a} \in R^{n \times 1}$, with ${\bf b}$ being the classifier of the attentionally-pooled feature. The second line illustrates that our formulation is in fact symmetric, where the final score can be seen as the inner product between {\em two} attentional heatmaps defined over all $n$ spatial locations.
Fig.~\ref{fig:nwarch-math} illustrates our approach.

{\bf Top-down attention:} To generate prediction for multiple classes, we replace the weight matrix from \eqref{eq:order2} with class-specific weights:
\begin{align}
 score_{order2}(X,k) =  Tr(X^TX W_k^T), \qquad \text{where} \qquad X \in R^{n \times f}, W_k \in R^ {f \times f} \label{eq:multiclass}
\end{align}
One could apply a similar derivation to produce class-specific vectors ${\bf a}_k$ and ${\bf b}_k$, each of them generating a class-specific attention map. Instead, we choose to distinctly model class-specific ``top-down'' attention~\cite{baluch2011mechanisms,zhou2016learning,ullman1984visual} from  bottom-up visual saliency that is class-agnostic~\cite{rutishauser2004bottom}. We do so by forcing one of the attention parameter vectors to be class-agnostic - e.g., $b_k = b$. This makes our final low-rank attentional model
\begin{align}
score_{attention}(X,k) = {\bf t}_k ^T {\bf h}, \qquad \text{where} \qquad {\bf t}_k = X{\bf a}_k, {\bf h} = X {\bf b}
\end{align}
equivalent to an inner product between top-down (class-specific) ${\bf t}_k$ and bottom-up (saliency-based) ${\bf h}$ attention maps. Our approach of combining top-down and botom-up attentional maps is reminiscent of biologically-motivated schemes that {\em modulate} saliency maps with top-down cues~\cite{navalpakkam2006integrated}. This suggests that our attentional model can also be implemented using a single, combined attention map defined over all $n$ spatial locations:
\begin{align}
 score_{attention}(X,k) = {\bf 1}^T {\bf c}_k, \qquad \text{where} \qquad {\bf c}_k =   {\bf t}_k \circ {\bf h}, 
\end{align}
\noindent where $\circ$ denotes element-wise multiplication and ${\bf 1}$ is defined as before. We visualize the combined, top-down, and bottom-up attention maps ${\bf c}_k, {\bf t}_k, {\bf h} \in R^{n \times 1}$ in our experimental results.

{\bf Average pooling (revisited):} The above derivation allows us to revisit our average pooling formulation from \eqref{eq:pool}, replacing weights ${\bf w}$ with class-specific weights ${\bf w}_k$ as follows:
\begin{align}
score_{top-down}(X,k) = {\bf 1}^T X {\bf w}_k =  {\bf 1}^T {\bf t}_k \quad \text{where} \quad {\bf t}_k = X {\bf w}_k \label{eq:top-down}
\end{align}
From this perspective, the above derivation gives the ability to generate top-down attentional maps from {\em existing} average-pooling networks. While similar observations have been pointed out before~\cite{zhou2016learning}, it naturally emerges as a special case of our bottom-up and top-down formulation of attention. 

\subsection{Network Architecture}
\label{sec:att-pool-bilinear-nw}

We now describe our network architecture to implement the attentional pooling
described above.
We start from a state of the art base architecture,
ResNet-101~\cite{He_16}.
It consists of a stack of `modules', each of
which contains multiple convolutional, pooling or identity mapping streams.
It finally generates a $n_1\times n_2\times f$ spatial feature map, which is average pooled 
to get a $f$-dimensional vector and is then classified using a linear classifier.

Our attention module plugs in at the last layer, after the spatial feature map.
As shown in Fig.~\ref{fig:nwarch} (Method 1), we predict a single channel bottom-up
saliency map of same spatial resolution as the last feature map, using a
linear classifier on top of it ($X{\bf b}$).
Similarly, we also generate the $n_1 \times n_2 \times K$ dimensional 
top-down attention map $X{\bf a}$, where $K$ is number of classes.
The two attention maps are multiplied and spatially averaged
to generate the $K$-dimensional output predictions ($(X{\bf a})^T(X{\bf b})$).
These operations are equivalent to first multiplying the 
features with saliency  ($X^{T}(X{\bf b})$)
and then passing through a classifier (${\bf a}(X^T(X{\bf b}))$).

{\bf Pose:} While this unconstrained attention module automatically learns
to focus on relevant parts and gives a sizable boost in accuracy, we take inspiration
from previous work~\cite{cheronICCV15} and use human pose keypoints to guide
the attention. As shown in Fig.~\ref{fig:nwarch} (Method 2), we use a two-layer
MLP on top of the last layer to predict a 17 channel heatmap.
The first 16 channels
correspond to human pose keypoints and incur a $l_2$ loss against labeled (or
detected, using~\cite{cao2017realtime}) pose) The final channel is used as an unconstrained bottom-up attention map, as before. We refer to this method as 
pose-regularized attention, and it can be thought of as a non-linear
extension of previous attention map.

 \section{Experiments}

{\bf \noindent Datasets:}
We experiment with three recent, large scale action recognition
datasets, across still images and videos, namely
MPII, HICO and HMDB51. {\bf MPII Human Pose Dataset}~\cite{pishchulin14gcpr} contains
15205 images 
labeled with up to 16 human body keypoints, and classified into one of 
393 action classes.
It is split into train, val (from authors of~\cite{gkioxari15rstar})
and test sets, with 8218, 6987 and 5708 images each.
We use the val set to compare with~\cite{gkioxari15rstar} and for ablative
analysis while the
final test results are obtained by
emailing our results to authors of~\cite{pishchulin14gcpr}.
The dataset is highly
imbalanced 
and the evaluation is performed using mean average
precision (mAP) to equally weight all classes.
{\bf HICO}~\cite{chao15hico} is 
a recently introduced dataset with labels
for 600 human object interactions (HOI) combining 117 actions
with 80 objects.
It contains 38116 training and 9658 test images, with
each image labeled with all the HOIs active
for that image (multi-label setting).
Like MPII, this dataset is also highly unbalanced and
evaluation is performed using mAP over classes.
Finally, to verify our method's applicability to video based action recognition,
we experiment with
a challenging trimmed action classification dataset,
{\bf HMDB51}~\cite{hmdb51}. It contains 6766 realistic and varied
video clips from 51 action classes.
Evaluation is performed using average classification accuracy
over three train/test splits from~\cite{THUMOS13}, each with 3570 train
and 1530 test videos. \\
{\noindent \bf Baselines:}
Throughout the following sections, we compare our approach
first to the standard base architecture, mostly ResNet-101~\cite{He_16},
without the
attention-weighted pooling.
Then we compare to other reported methods and previous state of
the art on the respective datasets.

\begin{figure}
    \centering
        \includegraphics[trim={0 1cm 0 0},clip,width=\linewidth]{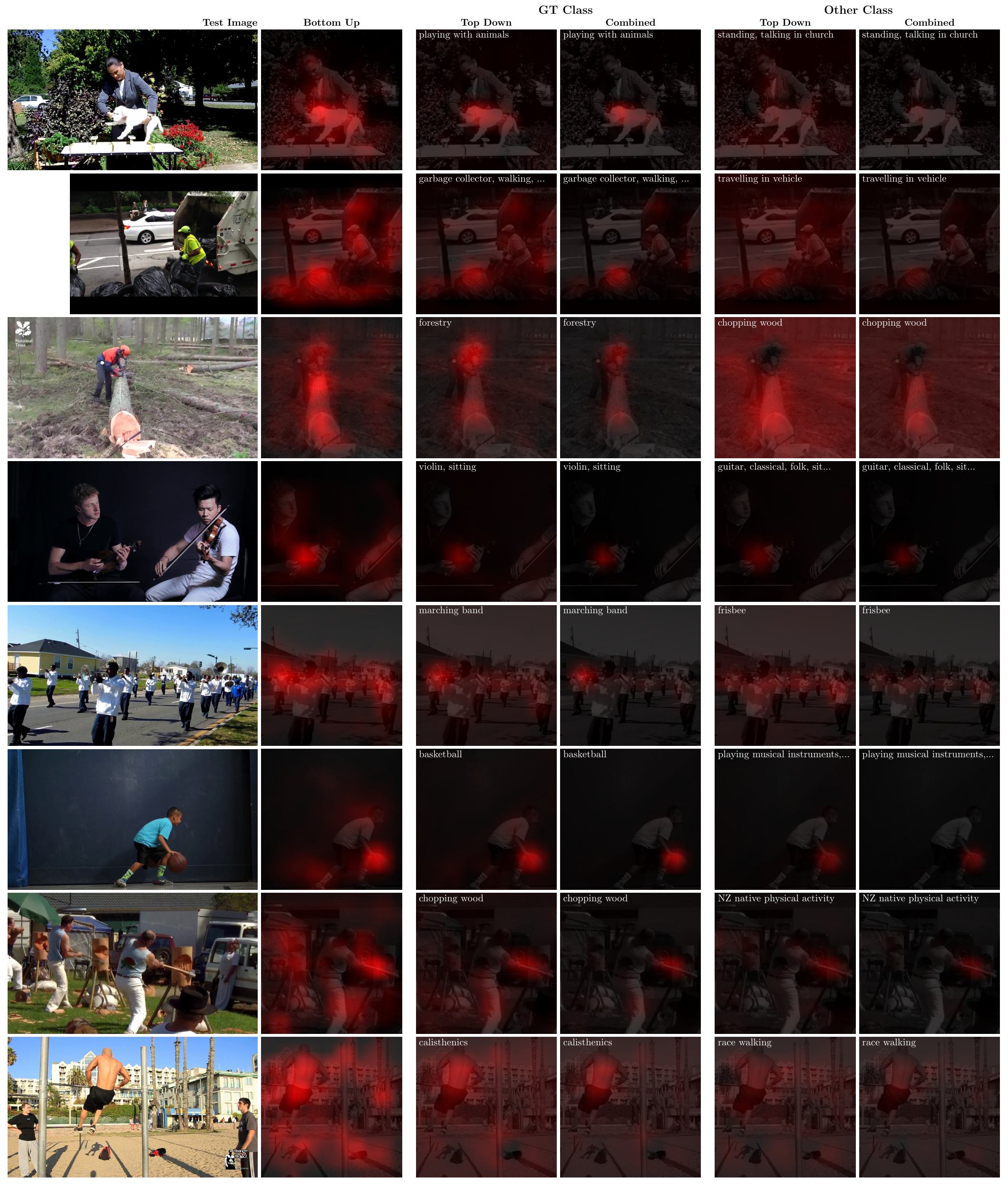}
    \caption{Auto-generated (not hand-picked) visualization of 
             bottom-up ($X{\bf b}$), top-down ($X{\bf a}_k$) and 
             combined ($(X{\bf a}_k) \circ (X{\bf b})$) attention on
             validation images in MPII, that see
             largest improvement in softmax score for correct class when trained
             with attention. Since the top-down/combined maps are class specific, we mention
             the class name for which they are generated for on top left of those heatmaps. We
             consider 2 classes, the ground truth (GT) for the image,
             and the class on which it gets lowest
             softmax score. The attention maps for GT class focus on the objects most
             useful for distinguishing the class. Though the top-down and combined maps look similar in many cases, they do capture
different information. For example, for a garbage collector action (second row), top-down also focuses on the vehicles in background, while the combined
map narrows focus down to the garbage bags. (Best viewed zoomed-in on screen)}
    \label{fig:mpii-heatmaps}
\end{figure}

{\noindent \bf MPII:}
We train our models for 393-way action classification on MPII with 
softmax cross-entropy loss for both the baseline ResNet and our 
attentional model.
We compare our performance in Tab.~\ref{tab:mpii}. Our unconstrained attention
model clearly out-performs the base ResNet model, as well as previous
state of the art methods involving detection of multiple
contextual bounding boxes~\cite{gkioxari15rstar} and 
fusion of full image with human bounding box
features~\cite{mallya16actions}.
Our pose-regularized model performs best,
though the improvement is small. We visualize the attention maps learned in
Fig.~\ref{fig:mpii-heatmaps}.

\begin{table}
\caption{Action classification performance on MPII dataset.
Validation (Val) performance is reported on train set split shared
by authors of~\cite{gkioxari15rstar}. Test performance
obtained from training on complete train set and submitting
our output file to authors of~\cite{pishchulin14gcpr}.
Note that even
though our pose regularized model uses pose labels at
training time for regularizing attention,
it does not require any pose input at test time. The {\bf top}-half corresponds to a diagnostic analysis of our approach with different base networks. Attention provides a strong 4\% improvement for baseline networks with larger spatial resolution (e.g., ResNet). Please see text for additional discussion. The {\bf bottom}-half reports prior work
that makes use of object bounding boxes/pose. Our method performs slightly better with pose annotations (on training data), but even without {\em any} pose or detection annotations, we outperform all prior work.}
\label{tab:mpii}
\tableSize{}
\begin{center}
\begin{tabular}{lccccrr}
\toprule
Method & Full Img & Bbox & Pose & MIL & Val (mAP) & Test (mAP) \\
\midrule
Inception-V2 (ours) & \checkmark  & & & & 25.2 & - \\
ResNet101 (ours) & \checkmark & & & & 26.2 & - \\
Attn. Pool. (I-V2) (ours) & \checkmark & & & & 24.3 & - \\
Attn. Pool. (R-101) (ours) & \checkmark & & & & {\bf 30.3} & {\bf 36.0} \\
\midrule
Dense Trajectory + Pose~\cite{pishchulin14gcpr} & \checkmark & & \checkmark & & - & 5.5 \\
VGG16, RCNN~\cite{gkioxari15rstar} & & \checkmark & & & 16.5 & - \\
VGG16, R*CNN~\cite{gkioxari15rstar} & & \checkmark & & & 21.7 & 26.7 \\
VGG16, Fusion (best)~\cite{mallya16actions} & \checkmark & \checkmark & & & - & 32.2 \\
VGG16, Fusion+MIL (best)~\cite{mallya16actions} & \checkmark & \checkmark & & \checkmark & - & 31.9 \\
Pose Reg. Attn. Pooling (R-101) (ours) & \checkmark & & \checkmark & & {\bf 30.6} & {\bf 36.1} \\
\bottomrule
\end{tabular}
\end{center}
\end{table}

\begin{figure}
    \centering
    \includegraphics[width=\linewidth]{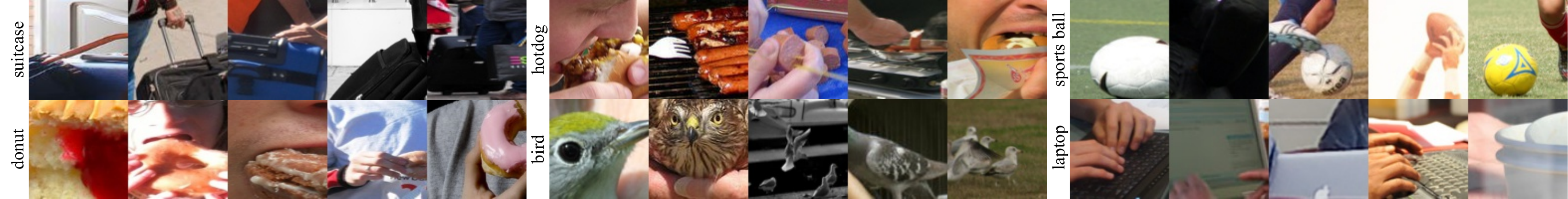}
    \caption{
We crop a 100px patch around the attention peak for all images containing an HOI
involving a given object, 
and show 5 {\bf randomly picked} patches for 6 object classes here. This suggests our
attention model learns to look for objects to improve HOI detection.
}
    \label{fig:hico-verb-object}
\end{figure}

{\noindent \bf HICO:}
We train our model on HICO similar to MPII, and 
compare our performance in Tab.~\ref{tab:hico}.
Again, we see a significant 5\% boost over our base ResNet model.
Moreover, we out-perform all previous methods, including ones that
use detection bounding boxes at test time except one~\cite{mallya16actions},
when that is 
trained with a specialized weighted loss for this dataset.
It is also worth noting that the full image-only performance
of VGG and ResNet were comparable in our experiments (29.4\% and 30.2\%),
suggesting that our approach shows larger relative improvement over
a similar starting baseline.
Though we did not experiment with the same optimization setting as~\cite{mallya16actions},
we believe it will give similar improvements there as well. 
Since this dataset also comes with labels decomposed into actions
and objects, we visualize what our attention model looks for, given
images containing interactions with a specific object. As Fig.~\ref{fig:hico-verb-object}
shows, the attention peak is typically close to the object of interest, showing
the importance of detecting objects in HOI detection tasks. Moreover, this suggests
that our attention maps can also function as weak-supervision for object detection.

\begin{table}
\vspace{-0.2in}
\caption{Multi-label HOI classification performance on HICO dataset.
The {\bf top}-half compares our performance to other full
image-based methods. The {\bf bottom}-half reports methods
that use object bounding boxes/pose. Our model out-performs
various approaches that need bounding boxes, multi-instance
learning (MIL) or specialized losses, and achieves
performance competitive to state of the art. Note that even
though our pose regularized model uses computed pose labels at
training time, it does not require any pose input at test time.
}\label{tab:hico}
\tableSize{}
\begin{center}
\begin{tabular}{lccccr}
\toprule
Method & Full Im. & Bbox/Pose & MIL & Wtd Loss & mAP \\
\midrule
AlexNet+SVM~\cite{chao15hico} & \checkmark & & & & 19.4 \\
VGG16, full image~\cite{mallya16actions} & \checkmark & & & & 29.4 \\
ResNet101, full image (ours) & \checkmark & & & & 30.2 \\
ResNet101 with CBP~\cite{gao16cbp} (impl. from~\cite{cbp_tf})
& \checkmark & & & & 26.8 \\
\methodTag{} (R-101) (ours) & \checkmark & & & & {\bf 35.0} \\
\midrule
R*CNN~\cite{gkioxari15rstar} (reported in~\cite{mallya16actions}) & & \checkmark & \checkmark & & 28.5 \\
Scene-RCNN~\cite{gkioxari15rstar} (reported in~\cite{mallya16actions}) & \checkmark & \checkmark & \checkmark & & 29.0 \\
Fusion (best reported)~\cite{mallya16actions} & \checkmark & \checkmark & \checkmark & & 33.8 \\
Pose Regularized \methodTag{} (R101) (ours) & \checkmark & \checkmark & & & {\bf 34.6} \\
\midrule
Fusion, weighted loss (best reported)~\cite{mallya16actions} & \checkmark & \checkmark & \checkmark & \checkmark & 36.1 \\
\bottomrule
\vspace{-0.3in}
\end{tabular}
\end{center}
\end{table}

\begin{figure}
    \vspace{-0.2in}
    \centering
    \includegraphics[width=\linewidth]{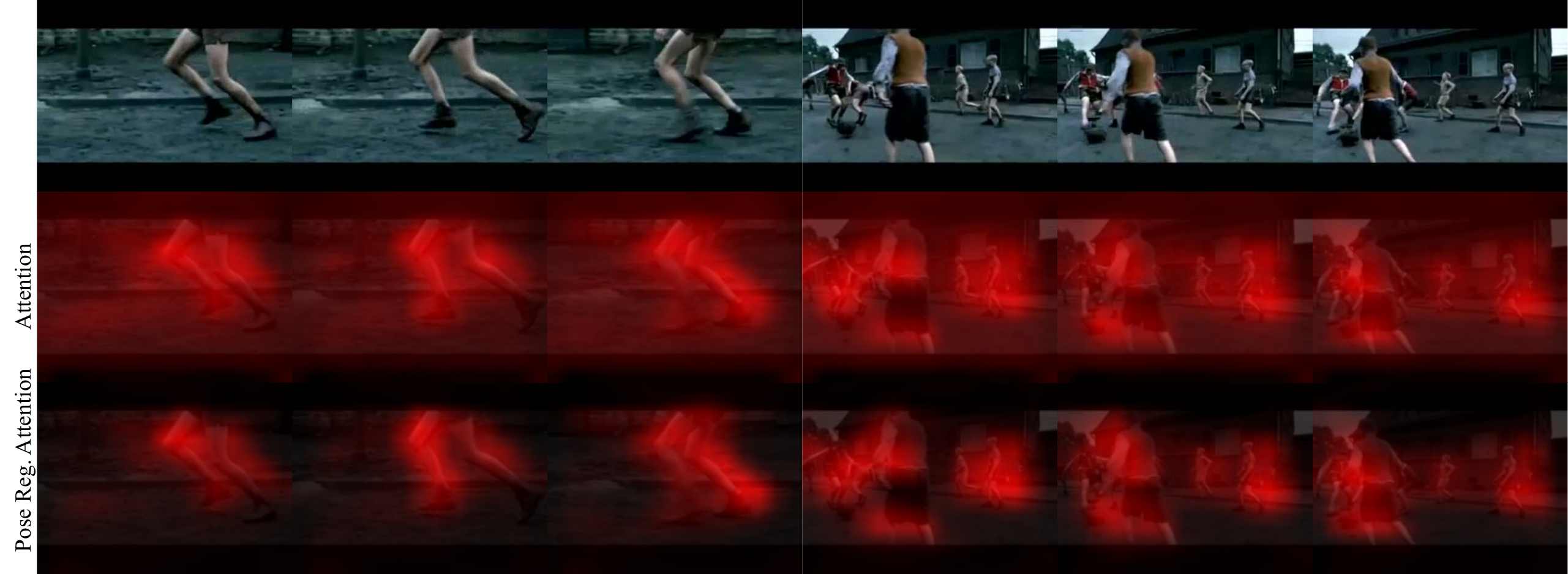}
    \caption{Attention maps with linear attention and pose regularized
attention on a video from HMDB. Note the pose-guided attention is better
able to focus on regions of interest in the non-iconic frames.}
    \label{fig:hmdb-att-maps}
\end{figure}

{\noindent \bf HMDB51:}
Next, we apply our attentional method to the RGB stream
of the current state of the art single-frame deep model on this
dataset, TSN~\cite{WangL_16a}. TSN extends the standard
two-stream~\cite{Simonyan_14b} architecture by using a much
deeper base architecture~\cite{Ioffe_15} along with enforcing
consensus over multiple frames from the video at training time.
For the purpose of this work, we focus on the RGB stream only
but our method is applicable to flow/warped-flow streams as well.
We first train a TSN model using ResNet-101 as base architecture
after re-sizing input frames to 450px. This ensures larger
spatial dimensions of the output ($14\times 14$), hence
ensuring the last-layer features are amenable to attention.
Though our base ResNet model does worse than BN-inception
TSN model, as Tab.~\ref{tab:hmdb} shows, using 
our attention module improves the
base model to do comparably well. Interestingly, on this dataset
regularizing the attention through pose gives a significant boost in
performance, out-performing TSN and establishing
new state of the art on the RGB-only single-frame model
for HMDB. We visualize the attention maps with normal and
pose-regularized attention in Fig.~\ref{fig:hmdb-att-maps}.
The pose regularized attention are more peaky near the human
than their linear counterparts. This potentially explains the
improvement using pose on HMDB while it does not help as much
on HICO or MPII; HICO and MPII, being image based datasets
typically have `iconic' images, with the subjects and objects of
action typically in the center and focus of the image. Video frames
in HMDB, on the other hand, may have the subject move all across
the frame throughout the video, and hence additional supervision
through pose at training time helps focus the attention at
the right spot.

\begin{table}
\caption{Action classification performance on HMDB51 dataset using only
the RGB stream of a two-stream model.
Our base ResNet stream training is done over 480px rescaled images,
same as used in our attention model for comparison purposes. Our
pose based attention model out-performs the base network by large
margin, 
as well as the previous RGB stream (single-frame) state-of-the-art, TSN~\cite{WangL_16a}.
}\label{tab:hmdb}
\tableSize{}
\begin{center}
\begin{tabular}{lrrrr}
\toprule
Method & Split 1 & Split 2 & Split 3 & Avg \\
\midrule
TSN, BN-inception (RGB)~\cite{WangL_16a}
(Via email with authors)
& {\bf 54.4} & 49.5 & 49.2 & 51.0 \\
ActionVLAD~\cite{Girdhar_17a_ActionVLAD} & 51.2 & - & - & 49.8 \\
RGB Stream, ResNet50 (RGB)~\cite{Feichtenhofer_16b}
(reported at~\cite{twofusion_web})
& - & - & - & 48.9 \\
RGB Stream, ResNet152 (RGB)~\cite{Feichtenhofer_16b}
(reported at~\cite{twofusion_web})
& - & - & - & 46.7 \\
TSN, ResNet101 (RGB) (ours) & 48.2 & 46.5 & 46.7 & 47.1 \\
Linear \methodTag{} (ours) & 51.1 & {\bf 51.6} & 49.7 & 50.8 \\
Pose regularized \methodTag{} (ours) & {\bf 54.4} & 51.1 & {\bf 50.9} & {\bf 52.2} \\
\bottomrule
\end{tabular}
\end{center}
\end{table}

{\noindent \bf Full-rank pooling:} Given our formulation of attention as low-rank second-order pooling, a natural question is what would be the performance of a full-rank model? Explicitly computing the second-order features of size $f\times f$ for $f=2048$ (and learning the associated classifier) is cumbersome. Instead, we make use of the compact bilinear approach (CBP) of~\cite{gao16cbp}, which generates a low-dimensional
approximation of full bilinear pooling~\cite{lin2015bilinear}
using the TensorSketch algorithm. To keep the final output comparable to our attentional-pooled model, we project to $f=2048$ dimensions. We find it performs slightly {\em worse} than simple average pooling in Table~\ref{tab:hico}.
Note that we use an existing implementation~\cite{cbp_tf} with minimal hyper-parameter optimization, and leave a more rigorous comparison to future work. \\
{\bf \noindent Rank-$P$ approximation:} While a full-rank model is cumbersome, we can still explore the effect of using a higher, $P$-rank approximation.
Essentially, a rank-$P$ approximation generates $P$ (1-channel) bottom-up and ($C$ channel) top-down attention maps, and the final prediction is the product of corresponding heatmaps, summed over $P$. On MPII, we obtain mAP of 30.3, 29.9, 30.0 for $P$=1, 2 and 5 respectively, showing that the validation performance is relatively stable with $P$. We do observe a drop in training loss with a higher $P$, indicating that a higher-rank approximation could be useful for harder
datasets and tasks. 
{\noindent \bf Per-class attention maps:} As we described in Sec.~\ref{sec:att-pool-bilinear},
our inspiration for combining class-specific and class-agnostic classifiers (i.e.\ top-down and bottom-up attention respectively),
came from the Neuroscience literature on integrating top-down and bottom-up
attention~\cite{navalpakkam2006integrated}.
However, our model can also be extended to learn completely class-specific attention maps, by predicting $C$ bottom-up attention maps,
and combining each map with the corresponding softmax classifier for that class.
We experiment with this idea on MPII and 
obtain a mAP of 27.9 with 393 (=num-classes) attention maps, compared to 30.3\% with 1 map, and 26.2\% without attention.
On further analysis we observe that both models achieve near perfect mAP on training data,
implying that adding more parameters with multiple attention maps leads to over-fitting on the relatively small
MPII trainset. However, this may be a viable approach for larger datasets.

{\noindent \bf Diagnostics:} It is natural to consider variants of our model that only consider the bottom-up or top-down attentional map. As derived in \eqref{eq:top-down}, baseline models with average pooling are equivalent to ``top-down-only'' attention models, which are resoundingly outperformed by our joint bottom-up and top-down model. It is not clear how to construct a bottom-up only model, since it is class-agnostic, making it difficult to produce class-specific scores.
Rather, a reasonable approximation might be applying an off-the-shelf (bottom-up) saliency method used to limit
the spatial region that features are averaged over. Our initial experiments with existing saliency-based methods~\cite{huang2015salicon} were not promising.

{\noindent \bf Base Network:} Finally, we analyze the choice of base architecture for
the effectiveness of our proposed attentional pooling module.
In Tab.~\ref{tab:mpii}, we compare the improvement using
attention over ResNet-101 (R-101)~\cite{He_16} and
an BN-Inception (I-V2)~\cite{Ioffe_15}. Both models
perform comparably when trained for full image,
however, while we see a
4\% improvement on R-101 on using
attention, we do not see similar improvements for I-V2.
This points to an important distinction in the two
architectures, i.e., Inception-style models
are designed to be faster 
in inference and training by rapidly down sampling input
images in initial layers through max-pooling. While this
reduces the computational cost for
later layers, it leads to most layers having very large
receptive fields, and hence later neurons have
effective access to all of the image pixels.
This suggests that
all the spatial features at the last layer could be highly
similar.
In contrast, R-101 downscales the spatial
resolution gradually, allowing the last layer features
to specialize to different parts of the image, hence
benefiting more from attentional pooling.
This effect was further corroborated by our experiments on
HMDB, where using the standard 224px input resolution showed
no improvement with attention, while the same image
resized to 450px at input time did. This initial
resize ensures the last-layer features are sufficiently
distinct to benefit from attentional pooling.

 \section{Discussion and Conclusion}

An important distinction of our model from some previous works~\cite{gkioxari15rstar,mallya16actions} is that it does not explicitly model action
at an instance or bounding-box level. This, in fact, is a strength of our model; making it
capable of attending to objects outside of any person-instance bounding box (such as bags of garbage for
``garbage collecting'', in Fig \ref{fig:mpii-heatmaps}). In theory, our model can also be applied to 
instance-level action recognition by applying attentional pooling over an instance's RoI features.
Such a model would learn to look at different parts of human body and its interactions with nearby objects.
However, it's notable that most existing action datasets, including~\cite{carreira2017quo,chao15hico,hmdb51,pishchulin14gcpr,charades,ucf101},
come with only frame or video level labels;
and though \cite{gkioxari15rstar,mallya16actions} are designed for instance-level recognition, they
are not applied as such. They either copy image level labels to instances or use multiple-instance learning,
either of which can be used in conjunction with our model.
Another interesting connection that emerges from our work is the relation between second-order pooling and attention.
The two communities are traditionally seen as distinct, and our work strongly suggests that they should mix:
as newer action datasets become more fine-grained, we should explore second-order pooling techniques for action recognition.
Similarly, second-order pooling can serve as a simple but strong baseline for the attention community, which tends to focus on more complex sequential attention networks (based on RNNs or LSTMs).
It is also worth noting that similar ideas involving self attention and bilinear models have recently also shown significant improvements in other tasks like 
image classification~\cite{wang2017residual}, language translation~\cite{vaswani2017attention} and visual question answering~\cite{santoro2017simple}. 
 \vspace{-0.1in}
\paragraph{Conclusion:}

We have introduced a simple formulation of attention as low-rank second-order pooling, and illustrate it on the task of action classification from single (RGB) images. Our formulation allows for explicit integration of bottom-up saliency and top-down attention, and can take advantage of additional supervision when needed (through pose labels). Our model produces competitive or state-of-the-art results on widely benchmarked datasets, by learning where to look when pooling features across an image.  Finally, it is easy to implement and requires few additional parameters, making it an attractive alternative to standard pooling, which is a ubiquitous operation in nearly all contemporary deep networks.

 \paragraph{Acknowledgements:} Authors would like to thank Olga Russakovsky for initial review. 
This research was supported in part by the National Science Foundation (NSF) under grant numbers CNS-1518865 and  IIS-1618903, and the Defense Advanced Research Projects Agency (DARPA) under Contract No. HR001117C0051. Additional support was provided by the Intel Science and Technology Center for Visual Cloud Systems (ISTC-VCS). Any opinions, findings, conclusions or recommendations expressed in this material are those of the authors and do not necessarily reflect the view(s) of their employers or the above-mentioned funding sources.
 
{\small
\bibliographystyle{abbrv}  \bibliography{refs}
}

\end{document}